\pdfoutput=1

\documentclass[11pt]{article}

\usepackage[final]{acl}

\usepackage{times}
\usepackage{latexsym}

\usepackage[T1]{fontenc}

\usepackage[utf8]{inputenc}

\usepackage{microtype}

\usepackage{inconsolata}

\usepackage{graphicx}
\usepackage{amsmath}
\usepackage{booktabs}
\usepackage{multirow}
\usepackage{float}
\usepackage{algorithm}
\usepackage{algpseudocode}
\usepackage{adjustbox}
\usepackage{subcaption}
\title{Transformers for Complex Query Answering over
Knowledge Hypergraphs}

\author{Hong Ting Tsang, Zihao Wang, Yangqiu Song \\
        CSE, HKUST, HKSAR, China \\ 
        \texttt{\{httsangaj,zwanggc,yqsong\}@cse.ust.hk}}
\begin{document}
\maketitle
\begin{abstract}
Complex Query Answering (CQA) has been extensively studied in recent years. In order to model data that is closer to real-world distribution, knowledge graphs with different modalities have been introduced. Triple KGs, as the classic KGs composed of entities and relations of binary, have limited representation of real-world facts. Real-world data is more sophisticated. While hyper-relational graphs have been introduced, there are limitations in representing relationships of varying arity that contain entities with equal contributions. To address this gap, we sampled new CQA datasets: JF17k-HCQA and M-FB15k-HCQA. Each dataset contains various query types that include logical operations such as projection, negation, conjunction, and disjunction. In order to answer knowledge hypergraph (KHG) existential first-order queries, we propose a two-stage transformer model, the Logical Knowledge Hypergraph Transformer (LKHGT), which consists of a Projection Encoder for atomic projection and a Logical Encoder for complex logical operations. Both encoders are equipped with Type Aware Bias (TAB) for capturing token interactions. Experimental results on CQA datasets show that LKHGT is a state-of-the-art CQA method over KHG and is able to generalize to out-of-distribution query types.
\end{abstract}

\section{Introduction}
Knowledge graphs (KG), comprising facts with entities as nodes and relations as edges, have gained much attention recently. Triplets (head, relation, tail) represent binary relational facts in KG. Under the Open World Assumption (OWA) \citep{jiSurveyKnowledgeGraphs2022}, most existing graphs are considered incomplete. Both embedding methods and message passing approaches have been proposed to learn the underlying representations of entities and relations in KG, aiming to mine undiscovered facts.

Complex Query Answering (CQA) in KG is an important task in Neural Graph Database (NGDB) \citep{renNeuralGraphReasoning2023b}. In NGDB, KG serves as the data source, where CQA involves performing logical queries over incomplete KG via parameterization of entities, relations, and logical operators like projection, negation, conjunction, and disjunction. Various neural query approaches have been developed to infer missing links in KG and enrich the answer set.

Existentially quantified First Order queries with a single variable (EFO-1) are the primary type of complex query. While classic KG representations \citep{trouillonComplexEmbeddingsSimple2016} excel in one-hop inference, they fall short for complex queries. Different approaches have been proposed for answering EFO-1 queries. Embedding models \citep{chenFuzzyLogicBased2022, renBetaEmbeddingsMultiHop2020, zhangConEConeEmbeddings2021} model entities and relations in vector spaces, transforming queries into operator tree formats, with logical operators interacting with entity embeddings. Other methods, like CQD \citep{arakelyanComplexQueryAnswering2021} and LMPNN \citep{wangLogicalMessagePassing2023c}, utilize pretrained KG embeddings to tackle EFO-1 queries. CQD formulates logical inference as an optimization problem, while LMPNN treats EFO-1 queries as query graphs, instantiating nodes with pretrained embeddings and executing message passing to predict answer embeddings.

However, less attention has been given to complex queries beyond binary relations. Most current models focus on ordinary KG, despite the abundance of n-ary facts in modern large KG. Hypergraphs and hyper-relation graphs are suitable for constructing KG to accommodate n-ary relations. StarQE \citep{alivanistosQueryEmbeddingHyperrelational2022a} introduces hyper-relational graphs in WD50K-QE, where each triple is built from qualified statements (head, relation, tail, qp), with qp as contextual information. However, hyper-relational graphs differ from hypergraphs; they treat n-ary facts as triples with supporting facts, focusing on query answering.

Hypergraphs better represent entities with equal contributions in relations. For instance, the relation 
coauthor(authorA,authorB,authorC) can be correctly depicted by a hyperedge in a hypergraph but misrepresented in a hyper-relational graph. Efforts have been made to create Knowledge Hypergraph (KHG) embeddings in M-FB15k \cite{fatemiKnowledgeHypergraphsPrediction2020a} and JF17k \citep{wenRepresentationEmbeddingKnowledge2016a}, primarily for link prediction. However, little work has focused on CQA in KHG, with LSGT \citep{baiUnderstandingInterSessionIntentions2024} being the only study investigating CQA with ordered hyperedges, using binary relations for intermediate variable nodes.

In this paper, we propose a novel two-stage transformer model, Logical Knowledge Hypergraph Transformer (LKHGT), a transformer-based approach for reasoning over KHGs, extending CQA from binary to n-ary relations. LKHGT consists of two encoders: the Projection Encoder, which predicts answers for variables in atomic hyperedges, and the Logical Encoder, which derives intermediate variable embeddings for logical operations. Both encoders are equipped with Type Aware Bias (TAB), which helps identify each input token type and capture their interactions. We define the logical query in EFO-1 Disjunctive Normal Form and construct ordered query hypergraphs. Each ordered hyperedge represents an atomic formula, including a predicate and potentially a negation operator.

We investigate LKHGT's performance against baseline models. To ensure fair experiments, we included multiple methods capable of reasoning on hypergraphs and extended LMPNN \citep{wangLogicalMessagePassing2023c} to relational ordered knowledge hypergraphs, utilizing pretrained KHG embeddings with logical inference capabilities.

Our experiments show that LKHGT outperforms other n-ary CQA models in ordered hyperedge settings. The model generalizes from edges of binary to arity N and achieves promising results. After learning basic logical operations, our model also demonstrates good performance on out-of-distribution queries. By comparing the Logical Encoder and fuzzy logic \citep{zadehFuzzyLogic1988} for logical operations, we show that our transformer-based encoder outperforms neural symbolic methods using fuzzy logic. Thus, our approach bridges the gap between CQA and Knowledge Hypergraphs.

\section{Related Works}

\subsection{Knowledge Hypergraphs Embedding}
BoxE \citep{abboudBoxEBoxEmbedding2020}, HypE, ReAlE \citep{fatemiKnowledgeHypergraphEmbedding2021a, fatemiKnowledgeHypergraphsPrediction2020a}, n-Tucker \citep{liuGeneralizingTensorDecomposition2020a}, m-TransH \citep{wenRepresentationEmbeddingKnowledge2016a} and RAE \citep{zhangScalableInstanceReconstruction2018} are Knowledge Hypergraphs Embeddings that considered n-ary facts in the form of r(e1, e2 . . .). These methods aims to perform link prediction by encoding entities and relations into embeddings continuous spaces. These models are equivalently performing projection on atomic hyperedge. Although they can perform well in atomic query projection task, they lack the ability to perform multi-hop query, where performance degrades as projection continues on sub-queries. To enable logical operations on these embeddings, fuzzy logic \citep{zadehFuzzyLogic1988, chenFuzzyLogicBased2022} or message passing methods \citep{wangLogicalMessagePassing2023c} can be applied.

\subsection{Complex Query Answering}
Complex Query Answering (CQA) leverages various approaches, including transformers, query embeddings, neural-symbolic methods, and message passing techniques. The introduction of Logical Message Passing Neural Networks (LMPNN) \citep{wangLogicalMessagePassing2023c} has facilitated the use of different pretrained Knowledge Graph (KG) embeddings, which serve as initial entity and relation embeddings. Final answer embeddings are obtained through message passing on query graphs, utilizing embeddings such as RESCAL \citep{nickelThreewayModelCollective2011}, TransE \citep{bordesTranslatingEmbeddingsModeling2013}, DistMult, ComplEx \citep{trouillonComplexEmbeddingsSimple2016}, ConvE \citep{dettmersConvolutional2DKnowledge2018}, and RotatE\citep{sunRotatEKnowledgeGraph2019a}.

For query embeddings, GQE \citep{hamiltonEmbeddingLogicalQueries2019a} addresses queries with existential quantifiers and conjunctions, while FuzzQE \citep{chenFuzzyLogicBased2022} employs fuzzy logic to define logical operators. In the neural-symbolic domain, methods like BetaE \citep{renBetaEmbeddingsMultiHop2020}, ConE \citep{zhangConEConeEmbeddings2021}, QUERY2BOX \citep{renQuery2boxReasoningKnowledge2020} and ENeSy \citep{xuNeuralSymbolicEntangledFramework2022}, project symbols into continuous spaces. MLPMix \citep{amayuelasNeuralMethodsLogical2022} and NewLook \citep{liuNeuralAnsweringLogicalQueries2021} utilize MLPs and attention mechanisms for CQA. Additionally, QTO \citep{baiAnsweringComplexLogical2023}, and CQD \citep{arakelyanComplexQueryAnswering2021} apply combinatorial optimization to query computation trees.

Since the introduction of Transformers \citep{vaswaniAttentionAllYou2023b}, several methods have applied this architecture to CQA. SQE \citep{baiSequentialQueryEncoding2023} linearizes EFO-1 queries for sequence encoding, PathFormer \citep{zhang_pathformer_2024} recursively process EFO-1 queries tree like NQE \citep{luoNQENaryQuery2023b} and Tree-LSTM did \citep{taiImprovedSemanticRepresentations2015}. While QTP \citep{xuQuery2TripleUnifiedQuery2023} separates simple and complex queries, employing distinct neural link predictors and query encoders.

Research on inductive biases includes relative positional encoding \citep{shawSelfAttentionRelativePosition2018} and rotary embedding \citep{suRoFormerEnhancedTransformer2023}, enhancing the self-attention mechanism. Additionally, HAN \citep{wangHeterogeneousGraphAttention2021} implements transformers for link prediction in knowledge graphs, and KnowFormer \citep{liuKnowFormerRevisitingTransformers2024} tailors the transformer for simple query answering. TEGA \citep{zhengEnhancingTransformersGeneralizable2025} applies inductive biases based on token type interactions for EFO-syntax queries.

Previous studies on CQA have primarily focused on knowledge graphs, with extensive research on ordinary knowledge graphs addressing ordinary query answering (OWA) by predicting unseen triples. EFO-1 queries can be effectively answered when the arity equals two. However, few attempts have been made to extend these approaches to general n-ary facts in hyperedge or hyper-relational formats.

\subsection{N-ary Graph Reasoning}
First N-ary reasoning problem introduced in the field of CQA is performed on hyper-relational knowledge graph. StarQE \citep{alivanistosQueryEmbeddingHyperrelational2022a} utilize StarE \citep{galkinMessagePassingHyperRelational2020} as graph encoder for StarQE, equip it with message passing to have the ability to deal with multi-hop queries. TransEQ \citep{liuGeneralizingHyperedgeExpansion2024} is an query embedding models that generalize star expansion \citep{agarwalHigherOrderLearning2006} for hyper-edges to hyper-relational graphs, then use encoder-decoder to capture structural information and semantic information for hyper-relational knowledge graph completion task. NeuInfer \citep{guanNeuInferKnowledgeInference2020} chose to represent n-ary fact as a primary triple coupled with a set of its auxiliary descriptive attribute-value pair(s) and use neural network to perform knowledge inference. NQE \citep{luoNQENaryQuery2023b} use dual-heterogeneous Transformer encoder and fuzzy logic \citep{zadehFuzzyLogic1988} to recursively process hyper-relational query tree. However, hyper-relational edges encode entities with possibly different relation in a single triples, which does not exhibit same characteristics as hyperedges. The encoder input format for NQE generalize n-ary inputs, thus it can be naturally extend to be used in hyper-edges query answering.
SessionCQA \citep{baiUnderstandingInterSessionIntentions2024}, The first CQA model incorporates the concept of hyper-edges, encoding user sessions as hyper-edges for item recommendations. However, its query type does not follow a fully hyper-edge setting; only the initial first hop is represented as a hyper-edge, while subsequent hops use binary relation edges.
The query input format for SessionCQA can also naturally extend for N-ary facts.

\section{PRELIMINARIES}
\subsection{Knowledge Hypergraphs}
 A knowledge hypergraph \( \mathcal{G} = (\mathcal{E}, \mathcal{R}, \mathcal{H}) \), where \( \mathcal{E} \) is the set of entities \( e \) in the knowledge hypergraph, \( \mathcal{R} \) is the set of relations \( r \), and \( \mathcal{H} \) is the set of ordered hyperedges \( h = r(e_1, \ldots, e_k) \in \mathcal{H} \), where \( e_1, \ldots, e_k \in \mathcal{E} \) and \( r \in \mathcal{R} \). The arity of a hyperedge \( h \) is defined as \( k = ar(r) \), where each relation type has a fixed arity size. Each position in a relation type has a specific semantic meaning that constructs the ordered hyperedge.
Although  through star expansion \citep{agarwalHigherOrderLearning2006} of hyperedge, an hyperedge can be converted to homogeneous graph, however structural information will be loss in process.

\subsection{EFO-1 Query in Hypergraphs}
\begin{figure*}  
    \centering
\includegraphics[width=\linewidth]{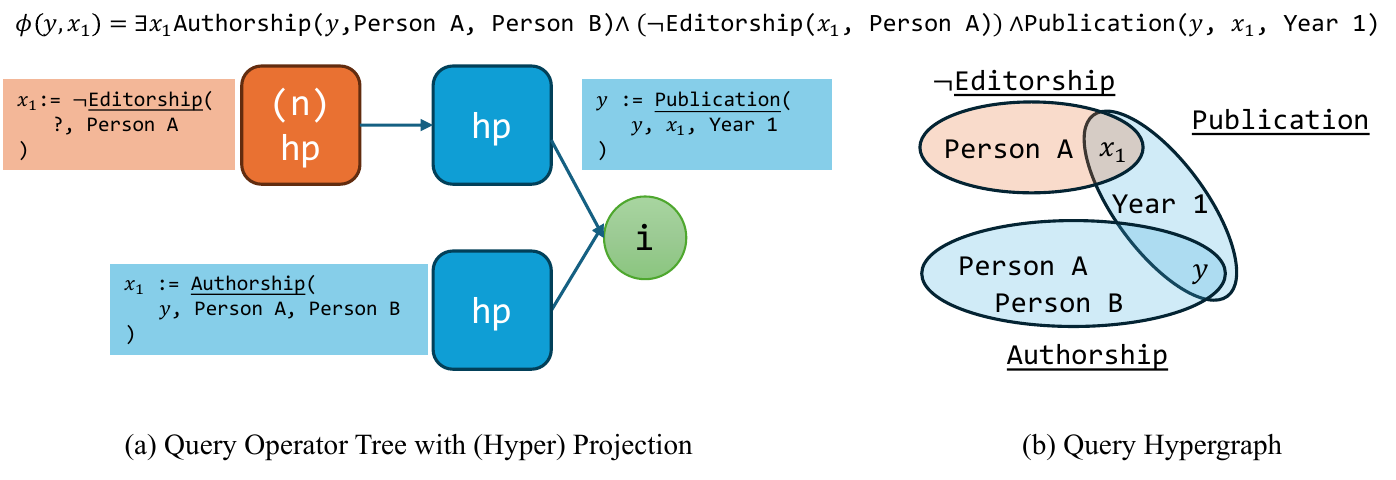}  
    \caption{Example of LKHGT processing query tree of ip type}
    \label{fig:example}
\end{figure*}
In this paper, we will focus on Existential First Order queries with a single free variable with logical formulas (EFO-1) under the disjunctive normal form in hypergraph settings. In the following discussions, we will refer to hyper-projection simply as projection without explicitly specifying it. We aim to emphasize the unique properties of projections within hypergraph structures. 

An atomic formula $a$ is composed of term, relation and variable. A simplest atomic formula can be intuitively represented by an ordered hyperedge $h = r_1(e_1,...,v_k)$, where $e_1$ is a term, $v_k$ is an variable at position k and $r_1$ is the relation. Let $ar(r)$ is the arity of the given relation $r$. A knowledge graph is a relational knowledge hypergraph where for all $r \in \mathcal{R}, ar(r) = 2$. An atomic formula can be negated by adding $\neg$ to form $\neg r(e_1,...,v_k)$. A first order formula can be iteratively constructed by atomic formula $a$ using connectives conjunction $\land$ and disjunction $\lor$. Quantifiers can be added to variables in $a$ using quantifiers like $\exists$ and $\forall$. Variables without quantifiers is considered as free.

Given a knowledge hypergraph $\mathcal{G}$, an EFO-1 query $q$ is defined as a first-order formula in the below Disjunctive Normal Form (DNF), 
\begin{multline}
q(y,x_1,...,x_n) =  \exists x_1, \ldots, \exists x_m [ (a_{11} \land a_{12} \land \cdots \\ \land a_{1n_1}) \lor \cdots 
\lor (a_{p1} \land a_{p2} \land \cdots \land a_{pn_p}) ]    
\end{multline}

where y is the only free variable, and $x_i$ for $1 \leq i \leq m$ are existential variables. Each brackets represent an complex query where each $a_{ij}$ is an atomic formula with constants $y,x_i$ and variables that can be either negated or not.
To address the EFO-1 queries, one must determine the answer set A[$Q,KG$]  such that for each a $\in$ A[$Q,KG$], $q(y=a, x_1,...,x_n)$ returns true.

Considering a query in the form of 
\begin{equation*}
    r_1(y_1,e_1,e_2) \land \neg r_2(x_1,e_3) \land r_3(y_1,x_1,e_4)
\end{equation*}
there are 2 ways to represent EFO-1 queries in hypergraphs, which are query graphs and operator tree.

\noindent \textbf{Operator tree.} In operator tree, each node itself is an operator node that corresponds to logical operations like, projection, conjunction, disjunction, negation. In knowledge hypergraph settings, each projection node consists of set of entities for itself and a variable nodes in random arity position. As Figure \ref{fig:example} shown, each atomic query is represented by the projection nodes. Then, two projection nodes pointed to the intersection node, and the intersection node represent the conjunction of answer sets from its pointers. Intersection node use the intersected variable to obtain the final answer.

\noindent \textbf{Query graph.} In Figure\ref{fig:example}, it shows an alternative way to represent EFO-1 query. Each hyper edge representing an atomic formula containing edge information like relation types and negation information. The nodes involved are either constant symbol, free variable or existential variable.

In this paper, each Knowledge Hypergraph EFO-1 query is presented in the form operator tree for LKHGT. As we can see that in order to allow transformer to perform message passing alike operation through the self-attention mechanism, we have to encode whole query in a single pass. When transformer based model encode whole query graphs as a sequence of tokens, it becomes more challenging for transformers to focus on solving each atomic query in the multihop-query. Thus, operator tree is more suitable for our choice of modeling.

\section{Logical Knowledge Hyper Graph Transformer}
\begin{figure*}  
    \centering
    \includegraphics[width=\linewidth]{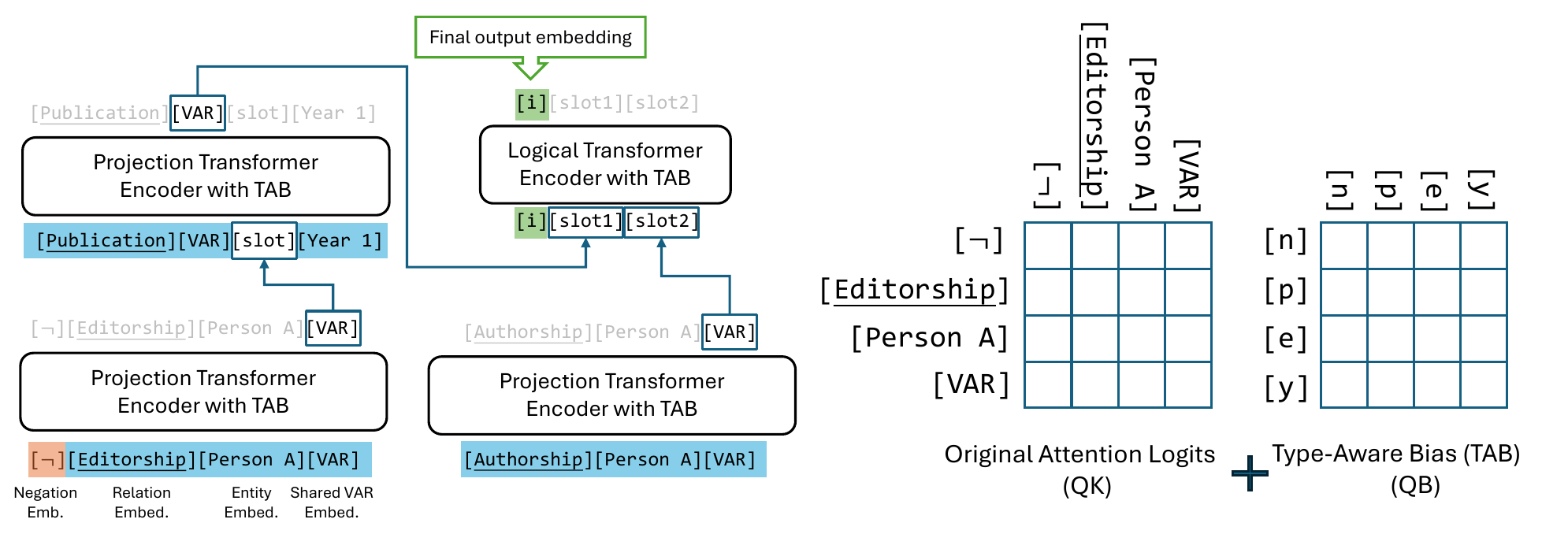}  
    \caption{Example of LKHGT processing query tree}
    \label{fig:example_op}
\end{figure*}
In this section, we will describe the how operator tree is used to represent complex query, and explain how the 2 stage encoder, Logical Knowledge Hyper Graph Transformer (LKHGT), works given a query tree input. In order to allow precise estimation of answer set, we propose to use 2 stage encoder to process the query tree iteratively. For example, refer to Figure \ref{fig:example_op}, we can see that each projection node at bottom level first receive necessary information like, negation, relation, entities and position of variable node inside the hyperedge. Then these information is fed into projection encoder for each projection. For second hop of the atomic query, previous projected variable embedding is fed as one of the input. Projection is performed using the answer set embeddings output from projection encoder. Finally, intersection is then performed with logical encoder, which is done by receive all variable embeddings output from each projection encoder. Although using iterative method to process query tree is slower comparing to encoding whole query graph at once, we maximize the correctness of answer set embeddings output for each atomic formula.
\subsection{Tokens Input and Output}
Operators type included in operator tree are projection (p), negation ($\neg$), intersection ($\land$), union ($\lor$). As defined, each complex query is composed of atomic formula which is made of relaiton (r), subject (s), variable (e). Overally speaking, there are 8 types of token. Relation, existential variable, free variable, entity and negation type of tokens will be fed into projection encoder. Projection, intersection, union types of token will be fed into logical encoder.

\noindent\textbf{Projection Encoder} Given the fact that negation can only applied to atomic formula which is represented as projection operation, we chose to process negation operation in projection encoder. Projection Encoder projects tokens into continuous space according to their token type. If negation is involved, the input tokens will be, 
\begin{equation}
    {\rm tokens} = T_p = [n,r_1,e_1,x_2,...] 
\end{equation}
The corresponding projected sequence will be,
\begin{equation}
    X = W_pT_p = [W_nn,W_rr_1,W_ee_1,...] \in R^{n \times d}
\end{equation}
where $W_{p} \in R^{d \times d} $is the linear projection weight, where $ p \in \{ n, r, x, e, y\} $.   
Since projection encoder is performing projection over ordered hyperedge, we will also add absolute positional encoding to entities node and variable node. Let the set of nodes be $N$ for an hyperedge, then the input embeddings will be.
\begin{equation}
    X_{v\in N} = X_{v\in N} + PE_{pos}
\end{equation}
\noindent\textbf{Logical Encoder}
Logical encoder will focus in dealing with logical task that involves more than 1 projection node, which is intersection (conjunction) and union (disjunction). These nodes will receive pointers from more than one operators node, we use transformer logical encoder to encode all projected variable, the input tokens for logical encoder will be,
\begin{equation}
    {\rm tokens} = T_l = [i/u,p_1,p_2,p_3,...] 
\end{equation}
The corresponding projected sequence will be,
\begin{equation}
    X = W_lT_l = [W_{i/u} \{i/u\},W_pp_1,...] \in R^{n \times d}
\end{equation}
where $W_{l} \in R^{d \times d} $is the linear projection weight, where $ l \in \{ i, u, p\} $. 
Both projected sequence is then further process with modified version of \citep{vaswaniAttentionAllYou2023b}. Finally, Projection Encoder will output embeddings of variable node in hyperedge and Logical Encoder will output the embeddings of the logical operator tokens.

\subsection{Type Aware Bias (TAB) for Projection Encoder and Logical Encoder}

Inspired by the idea of introducing inductive bias for meta paths in \citep{wangHeterogeneousGraphAttention2021}, LKHGT also introduced modified inductive bias tailored for CQA task. With the given 8 token types, we can observe there are types of token interaction bias to be added into self attention matrix. For Projection Encoder there are $P^5_2$ edge types for token types [$n,r,x,e,y$]. For Logical Encoder, there are $P^3_2$ edge types for token types. In order to cater for these edge types differences, we construct the attention matrix with Type Aware Bias (TAB) as below: 
\begin{equation}
    e_{ij} = \frac{x_i W^Q_o(x_j W^k_o + B_{ij})^T}{\sqrt{d}}, a_{ij} = {\rm softmax}(e_{ij})
\end{equation}
\begin{equation}
    z_i = \sum^n_{j=1} a_{ij}(x_j W^V_o)
\end{equation}
Where ($W^Q,W^K,W^V$) is the linear projection matrix $\in R^{d\times d}$ and $\{ o, i, j\}$ can be the eight token types $ t \in \{n,r,x,e,y,i,u,p\}$, $B_{ij}$ is the Type Aware Bias (TAB) vector in token interaction bias matrix $B \in R^{|t|\times |t| \times d}$. These bias able to differentiate the token types interaction and help facilitates the capture of nuanced interactions among token types, allowing the transformer to adaptively weigh the importance of each token based on its role in the context. 

In our case, we can treat Transformer as a special case of GNN. For Projection encoder, each hyperedge is an fully connected graph input in the perspective of GNN. We are doing aggregation with attention on this fully connected graph and predict the embeddings of the projection node using transformer self-attention mechanism. This aggregation process not only enriches the embeddings of the projection node but also ensures that the model can dynamically adjust to the significance of different tokens based on the context. The result is a robust embedding that encapsulates the rich relational structure inherent in the input data.

\subsection{Transformer as the replacement for Fuzzy Logic}
Transformers present a compelling replacement for fuzzy logic in current CQA models, primarily due to their enhanced representation capabilities and scalability. While fuzzy logic effectively manages uncertainty and ambiguity, Transformers leverage attention mechanisms to generate richer and more nuanced embeddings. Learning from vast amounts of data rather than relying on predefined rules, allowing for a better contextual understanding of projected nodes' embeddings. While fuzzy logic has its strengths, Transformers provide a more powerful and adaptable framework for processing and representing information in CQA models.

\subsection{Training LKHGT}
After obtaining the output embeddings $f \in R^{d}$, we can compute the output label of the entity by a simple MLP decoder and have output $\hat{y} \in R^{|\mathcal{E}|}$. Given a the query embedding $\hat{y}$, we can first use a softmax function to obtain the probability scores of query embedding between all entity,
\begin{equation}
    \sigma (\hat{y}) = \frac{e^{\hat{y}_i}}{\sum_j e^{\hat{y}_j}}
\end{equation}
Then we can construct the cross-entropy loss to maximize the log probability of $\hat{y}$ matching the answer $a$,
\begin{equation}
    L = - \frac{1}{N} \sum_i \log (\sigma(\hat{y}))
\end{equation}
where N is the batch size.
\begin{table*}
\scriptsize
\caption{MRR results of different CQA models on two KHGs. AP represents the average score of EPFO queries and AN represents the average score of queries with negation. The boldface indicates
the best results for each KG.}
\centering
\resizebox{\textwidth}{!}{
\begin{tabular}{llllllllllllllllll} \toprule
KHG & Model & 1P & 2P & 3P & 2I & 3I & PI & IP & 2U & UP & 2IN & 3IN & INP & PIN & PNI & AP & AN \\ \midrule
\multirow{ 2}{*}{JF17k}  
 & HLMPNN & 45.66 &  6.58 & 12.24 & 46.03 & 47.27 & 15.52 &  7.91 & 25.45 & 12.75 & 10.43 & 14.49 & 10.83 &  7.26 &  8.29 &      24.38 &     10.26 \\
 & LSGT & 49.51 & 21.73 & 13.10 & 50.46 & 34.85 & 32.07 & 23.00 & 22.27 & 20.33 & 16.05 & 13.45 &  8.44 & 10.10 & 12.39 &      29.70 &     12.09 \\ 
 & NQE &  56.19 & 38.29 & 34.72 & 64.88 & \textbf{70.40} & 53.95 & 38.24 & 29.11 & 38.39 & 22.51 & \textbf{27.80} & 15.04 & 17.84 & 21.83 &      47.13 &     21.00\\ 
 & LKHGT &\textbf{58.19} & \textbf{41.87} & \textbf{38.76} & \textbf{68.47} & 42.71 & \textbf{57.48} & \textbf{42.01} & \textbf{33.38} & \textbf{41.70} &\textbf{26.28} & 18.26 & \textbf{20.66} & \textbf{18.32} & \textbf{24.87} &      \textbf{47.17} &     \textbf{21.68} \\ 

 \hline
\multirow{ 2}{*}{M-FB15k}  
 & HLMPNN & \textbf{47.86} & 28.04 & 23.02 & \textbf{45.78} & 45.65 & 32.89 & 26.12 & 25.35 & 26.27 & 17.32 & 22.58 & 21.98 &  9.47 & 15.85 &      33.44 &     17.44 \\
 & LSGT & 44.15 & 30.35 &  9.87 & 39.47 & 31.19 & 30.01 & 28.57 & 16.46 & 27.28 & 16.55 & 10.10 & 10.59 & 11.37 & 19.29 &      28.59 &     13.58 \\ 
 & NQE & 46.43 & 33.43 & 29.80 & 43.75 & \textbf{48.84} & 34.94 & 32.09 & 16.83 & 30.20 & 22.26 & \textbf{26.28} & 20.19 & \textbf{17.51} & 24.95 &      35.15 &     \textbf{22.24} \\ 
 & LKHGT & 46.81 &\textbf{35.04} & \textbf{30.66} & 45.38 & 40.31 & \textbf{38.65} & \textbf{32.63} &\textbf{ 27.14} & \textbf{31.37} & \textbf{22.85} & 19.65 & \textbf{26.50} & 16.21 & \textbf{24.96} &      \textbf{36.44} &    22.03 \\ 
   
\bottomrule
    \end{tabular}}
\end{table*}

\begin{table*}[ht]
\scriptsize
\caption{MRR results for LKHGT in fuzzy settings, excluding absolute positional encoding and under a variable cardinality configuration.}
\centering
\resizebox{\textwidth}{!}{
\begin{tabular}{llllllllllllllllll} \toprule
KHG & Model & 1P & 2P & 3P & 2I & 3I & PI & IP & 2U & UP & 2IN & 3IN & INP & PIN & PNI & AP & AN \\ \midrule
\multirow{ 2}{*}{JF17k}  
 & LKHGT &  \textbf{58.19} & \textbf{41.87} & \textbf{38.76} &\textbf{ 68.47} & 42.71 & \textbf{57.48} & \textbf{42.01} & \textbf{33.38} & \textbf{41.70} & \textbf{26.28} & 18.26 & 20.66 & 18.32 &\textbf{ 24.87} &      47.17 &     21.68 \\
  & LKHGT w/ fuzzy. & 57.26 & 40.62 & 37.95 & 65.48 & \textbf{71.04} & 54.46 & 40.25 & 29.87 & 40.51 & 23.00 & \textbf{27.58} & 18.56 & \textbf{19.65} & 22.80 &      48.60 &     22.32 \\ 
 & LKHGT w/o abs. & 57.40 & 41.01 & 37.51 & 67.87 & 41.94 & 57.02 & 41.63 & 31.56 & 40.72 & 26.19 & 17.75 & \textbf{20.92} & 18.53 & 25.00 &      46.30 &     21.68 \\
 & LKHGT w/ var cardinality & 57.55 & 41.31 & 38.35 & 67.13 & 41.61 & 56.68 & 41.23 & 32.02 & 40.79 & 25.79 & 17.69 & 21.04 & 18.58 & 24.81 &      46.30 &     21.59 \\
\midrule
   & LKHGT w/ full training set & 58.43 & 43.89 & 42.75 & 69.11 & 75.19 & 58.19 & 43.16 & 32.05 & 42.51 & 26.26 & 31.62 & 23.18 & 18.74 & 25.35 &      51.70 &    25.03  \\
   & LKHGT w/ full train \& var card. & 58.54 & 44.40 & 43.58 & 69.20 & 74.46 & 58.56 & 43.56 & 32.91 & 42.90 & 26.48 & 31.51 & 24.06 & 19.30 & 25.59 &      52.01 &     25.39 \\ 
   
\bottomrule
\end{tabular}}
\end{table*}

\begin{table*}[ht]
\scriptsize
\caption{MRR results for CQA models under full training set.}
\centering
\resizebox{\textwidth}{!}{
\begin{tabular}{llllllllllllllllll} \toprule
KHG & Model & 1P & 2P & 3P & 2I & 3I & PI & IP & 2U & UP & 2IN & 3IN & INP & PIN & PNI & AP & AN \\ \midrule
\multirow{ 2}{*}{JF17k}  
 & LKHGT & 58.43 & 43.89 & 42.75 & 69.11 & 75.19 & 58.19 & 43.16 & 32.05 & 42.51 & 26.26 & 31.62 & 23.18 & 18.74 & 25.35 &      \textbf{51.70} &     \textbf{25.03} \\
 & LSGT &  50.56 & 21.85 & 17.75 & 53.34 & 53.71 & 32.06 & 20.78 & 22.49 & 18.33 & 14.00 & 16.40 &  6.05 &  7.44 &  8.59 &      32.32 &     10.50 \\ 
 & HLMPNN & 45.26 &  7.26 & 13.38 & 46.08 & 48.36 & 16.01 &  8.54 & 24.99 & 13.28 & 10.35 & 15.49 & 11.48 &  7.32 &  8.24 &      24.80 &     10.58 \\
 & NQE & 56.37 & 39.18 & 39.21 & 65.41 & 72.37 & 55.36 & 39.22 & 29.05 & 39.36 & 21.68 & 28.83 & 20.25 & 16.21 & 22.13 &      48.39 &     21.82 \\ 
   
\bottomrule
\end{tabular}}
\end{table*}
\begin{table}[ht]
\small
\caption{Statistics for each query types}
\centering
\begin{tabular}{llllllllllllllllll} \toprule
 & Dataset & 1P & Others \\ \midrule
 & Train & 60,000 & 20,000 \\
 & Valid &  10,000 & 10,000 \\ 
 & Test & 10,000 & 10,000 \\
\bottomrule
\end{tabular}
\end{table}
\section{Experiments}
In this section, we will describe the experiment set up including dataset, baselines and evaluation methods.
\subsection{Dataset}
Since Knowledge Hypergraph CQA is a subfield of CQA problem that has not been investigated before, thus we created custom Dataset JF17k-HCQA and M-FB15k-HCQA. Following the rationale from binary CQA sampling methods \citep{wangBenchmarkingCombinatorialGeneralizability2021a}, we modified the sampling methods into hypergraph version (Appendix \ref{sec:appendixC}), created datasets consists of query types (1P 2P 3P 2I 3I PI IP 2U UP 2IN 3IN INP PIN PNI), total 14 types, including all logical operation. Statistics for the number of examples sampled for each query type are listed in Table 4.

\subsection{Baselines}
For baseline models, we chose NQE \citep{luoNQENaryQuery2023b} and LSGT \citep{baiUnderstandingInterSessionIntentions2024} as our baseline models. NQE is a method that encode hyper-relational facts from HKG into embeddings. LSGT encode query graph information like, nodes ids, graph structure and logical structure using transformer.As NQE \citep{luoNQENaryQuery2023b} uses hyper-relational edges as input for its encoder. So, we followed the implementation of NQE in \citep{baiUnderstandingInterSessionIntentions2024} and modified our data format to feed it into NQE. For LSGT, we transformed our data into unified id format for input. For fair comparison, we replace all models encoder with simple basic transformer layer without modifying the other implementation details. Transformer embedding size is set to be 400 for iterative model like NQE and LKHGT. For LSGT, its embedding size is set to be 1024 in order to have it converge.
Aside from transformer based model, we also implemented HLMPNN (Appendix \ref{sec:appendixA}), a hypergraph version of LMPNN \citep{wangLogicalMessagePassing2023c}, which is set to have embedding size of 200. All models are trained on single 3090 GPU with 400 epochs.
\subsection{Evaluation}
In all experiments, we follow the common practice, sampled query from Knowledge Hypergrpahs $\mathcal{G}_{train}, \mathcal{G}_{val}, \mathcal{G}_{test}$, where $\mathcal{E}_{train} \subset \mathcal{E}_{val}$ and $\mathcal{E}_{val} \subset \mathcal{E}_{test}$. For all models, we will train on all query types except [3p, 3in, 3i, inp], in order to test the generalization of logical operations. We select the mean reciprocal rank (MRR),
\begin{equation}
    MRR(q) = \frac{1}{||v||}\sum_{v_i \in v} \frac{1}{rank(v_i)}
\end{equation}
as the evaluation metric to evaluate the ranking. For each query instance, we initially rank all entities, excluding those identified as easy answers, by their cosine similarity to the query variable embedding. The rankings of the hard answers are then used to calculate the Mean Reciprocal Rank (MRR) for that particular query instance. Finally, we compute the average of the metrics across all query instances. In this paper, we report and compare the MRR results.

\subsection{Results}
Table 1 presents the MRR results of LKHGT and other baseline models for answering EFO-1 queries across the two CQA datasets. It demonstrates that LKHGT achieves state-of-the-art performance on average for both EPFO and negation queries for n-ary hypergraph queries. When comparing LKHGT with transformer-based and message-passing methods, we observe that the message-passing methods underperform in complex query types. Additionally, when comparing the Iterative Model (LKHGT \& NQE) to the Sequential Model (LSGT), the Iterative Model shows superior performance. This disadvantage may be attributed to their approach of encoding the entire query at once, which can hinder performance on complex queries. In contrast, the LKHGT model shows notable improvements, particularly when replacing fuzzy logic with a logical encoder in the NQE framework, enhancing its ability to handle conjunctions and disjunctions effectively. Interestingly, LKHGT outperforms NQE on negation queries, despite incorporating negation in the Projection Encoder rather than relying on fuzzy logic. This suggests that fuzzy logic may not be sufficient for addressing complex query-answering tasks. The 1p, 2p, and 3p performance of LKHGT outperforming NQE is evidence that Type Aware Bias is contributing, as these query types do not involve the use of logical encoder. This distinction from the NQE model suggests that this bias is better at capturing the different interactions of token inputs. Overall, these findings confirm that the two-stage architecture of LKHGT successfully enhances performance in complex query scenarios, demonstrating its effectiveness and robustness in the context of query answering. \\
However, based on the performance of the LKHGT model in 3i and 3in, we observe that it is significantly lower than NQE. This discrepancy arises because the logical encoder did not encounter the input combinations generated by the output embedding of the projection encoder and the logical encoder itself, preventing it from effectively process the input. To provide a fair comparison of the effectiveness of LKHGT, we trained another instance using the full training set. The results are presented in Tables 2 and 3, where its performance in 3i and 3in shows significant improvement.
\subsection{Ablation Study}
In the ablation study, we conduct further experiments to justify the effects of the logical encoder and the positional encoding of LKHGT. We also investigate the effect of cardinality of input in logical encoder, where process embeddings a pair a time or process variable number of embeddings in a single run for operation like intersection and union. All experiments are performed on queries over JF17k with same setttings as before. Table 2 presents the results of the ablation study. From the results, we observe that without the positional encoding, the model's performance degrades, indicating the importance of this bias for identifying the position of each nodes to enhance performance. This may be due to the fact that each position contains a different semantic meaning within an hyperedge of ordered knowledge hypergraph. Such position information helps identify the entity properties in hyperedges. When using LKHGT with fuzzy logic, the results are still slightly better than NQE except for 3i and 3in due to the reason stated before, suggesting that the transformer-based projection encoder improve performance in negation and projection, and for logical encoder, it able to replace fuzzy logic operation, given that all types of input combination has been trained beforehand. For the Operator Cardinality, interestingly, we observe that for the transformer-based logical encoder, performing logical operations on a pair of inputs at a time yields better performance than processing all at once.
\section{Conclusion}
In this paper, we present the Knowledge Hypergraph CQA dataset to bridge the current gap in complex question answering (CQA) within hypergraph settings. We propose LKHGT to answer complex queries over Knowledge Hypergraphs, particularly EFO-1 queries. LKHGT achieves strong performance through its two-stage architecture with Type Aware Bias (TAB). In the ablation study, we demonstrate that the use of absolute positional encoding further enhances performance. Our research effectively addresses the gap in EFO-1 query answering tasks in hypergraph settings while emphasizing the advantages of transformer-based approaches for logical operations. The experiments show that, given sufficient training on various query types, LKHGT is the state-of-the-art model for current CQA tasks involving Knowledge Hypergraphs.
\section{Limitation}
A key limitation of our model is the time complexity linked to the projection and logical encoder components. The shift from fuzzy logic to a transformer-based logical encoder has increased complexity due to a larger parameter count and attention mechanisms. Additionally, the inductive bias requires training all combinations of token interactions, which can introduce noise if not properly managed, resulting in suboptimal performance and longer training times.

\noindent \textbf{Potential Risks}
The increased time complexity may lead to longer training and inference times, limiting practical applications. If the inductive bias is not addressed effectively, it might mislead the model toward noise rather than meaningful patterns. Furthermore, reliance on transformer architectures may create scalability challenges when dealing with larger datasets or complex tasks. Addressing these limitations and risks is crucial for the model's effectiveness.

\bibliography{ref}
\newpage
\appendix

\label{sec:appendix}

\begin{center}
    \Large Supplimentary Materials
\end{center}

\section{Appendix A} \label{sec:appendixA}
\subsection{Inspiration from LMPNN}
 Many previous works emphasize hypergraph message passing \citep{agarwalHigherOrderLearning2006, yadatiHyperGCNNewMethod2019, yadatiNeuralMessagePassing2020a}, with some tailored for hyper-relational graphs. Notable attempts include StarE \citep{galkinMessagePassingHyperRelational2020} and GRAN \citep{wangLinkPredictionNary2021}, which address one-hop queries on hyper-relational KGs. HR-MPNN \citep{huangLinkPredictionRelational2024a} established the General Relational Ordered Hypergraphs Message Passing Framework. Our baseline combines HR-MPNN \citep{huangLinkPredictionRelational2024a} with pretrained embeddings, drawing inspiration from LMPNN \citep{wangLogicalMessagePassing2023c}.
\subsection{A Natrual Extension of Logical Message Passing Nueral Network to Answer Hypergraph Queries}
\begin{itemize}
\item \textbf{Statement 1}: Prove $\rho(\{e_i | (e,i) \in N_j(h)\},r,j,0)$ is a generalization of $ \rho(h,r,h2t,0)$. \\
Where $N_i(h)$ as the positional neighborhood of a hyperedge $h$.\\
\\
Let $\rho$ be the hypergraph logical mesage encoding function of four input parameters, including neighboring entity embedding, relation embedding, query node position, and logical negation information.

Suppose that each edge with head at position 0 and tail at position 1, in the format of $(h,0)\rightarrow (t,1)$ for edge with arity = 2. Then with the definition of the ordered hyperedge,
there are 2 cases representing normal binary edge in hyperedge message encoding function. Their equivalent representation according to our definition:
\begin{enumerate}
    \item $ \rho(h,r,h2t,0)$ = $\rho(h,r, 1, 0)$
    \item $ \rho(t,r,t2h,0)$ = $\rho(t,r, 0, 0)$
\end{enumerate}
We simply replaced h2t and t2h flag with positional information.\\
For example, suppose a particular edge with arity = 6, in the form of $r(e_1,e_2,e_3,e_4,e_5,x)$, we can use the message encoding function express in the form of $\rho(\{e_1,e_2,e_3,e_4,e_5\},r,6,0)$.\\

\item \textbf{Statement 2}: Suppose $\rho(e_0,r,1,0) = \rho(h,r,h2t,0)$ and $\rho(h,r,h2t,0) = f(h,r)$ are true, where f is a binary function (e.g. elementwise multiplication).\\
Prove $\rho(\{e_i | (e,i) \in N_j(e)\},r,j,0) = f(g(\{N_j(e)\}),r)$ and is a generalization of KG message encoding function, where $g(\{N_j(e)\}) = f(f(e_1,...(f(e_{n-1},e_n))))$ is a function that recursively apply $f$.\\

As in LMPNN \citep{wangLogicalMessagePassing2023c}, there are 2 types of KG embeddings, characterized by their scoring functions, which are the inner-product-based scoring function and distance-based scoring function. The proof for closed-form logical messages for KHG representation is the same. Due to the fact that the properties of recursive function $g$ does not affect the proving in LMPNN\citep{wangLogicalMessagePassing2023c}, so as long as we can we prove our recursive function $g$ can obtain the same equation for normal KG closed-form logical messages with arity = 2, we can prove that: \\
\begin{equation*}
   \rho(\{e_i | (e,i) \in N_j(h)\},r,j,0) = f(g(\{N_j(e)\}),r)
\end{equation*}
Suppose we have an hyperedge $q$ and $j$ = 1, with arity = 2.
\begin{equation*}
N_j(q) = h 
\end{equation*} 
\begin{equation*}
g(N_j(q)) = f(h,r) 
\end{equation*} 
\begin{equation*}
    \rho(h,r,h2t,0) = \rho(\{e_i \mid (e,i) \in N_j(e_0)\}, r, j, 0) 
\end{equation*} 
\begin{equation*}
    \rho(\{e_i \mid (e,i) \in N_j(e_0)\}, r, j, 0) = f(h,r) 
\end{equation*} 

So for any edge with arity $\geq k$\\
\begin{equation*}
    \rho(\{e_i \mid (e,i) \in N_j(e_0)\}, r, j, 0) = f(g(N_j(e),r)
\end{equation*}

\end{itemize}

\subsection{Closed-form logical messages for HKG representation}
Table 5 is a table of KHG embeddings that can be express in the form of $\rho(\{e_i | (e,i) \in N_j(e)\},r,j,0)$
\begin{table*}
        \centering
        \begin{tabular}{@{}llcl@{}}
        \toprule
        \textbf{KHG Embedding}  & \textbf{f(h, r)} & \textbf{Estimate Function} \\ \midrule
        HypE        & $e_i \otimes e_j$          & $\otimes(r, g(\{N_j(e)\}))$ \\
        m-DistMult  & $e_i \otimes e_j$           & $\otimes(r, g(\{N_j(e)\}))$ \\
        m-CP        & $e_i \otimes e_j$           & $\otimes(r, g(\{N_j(e)\}))$ \\
        HSimplE     & $e_i \otimes \text{shift}(e_j,\text{len}(e_j)/\alpha) $                  & $\otimes(r, g(\{N_j(e)\}))$ \\ \bottomrule
        \end{tabular}
    \label{tab:kg_embeddings}
        \caption{Closed form foward estimation function $f$ for KHG representations.}
\end{table*}


\section{Sampling Algorithm for Knowledge Hypergraph Query} \label{sec:appendixC}
In this section, we introduce the algorithm for sampling EFO-1 queries from a Knowledge Graph of any arity, detailed in Algorithm 1. We adopt the general sampling approach for knowledge graphs from \cite{wangBenchmarkingCombinatorialGeneralizability2021a}. Given a graph \textit{G} and query type \textit{t}, we randomly select a node as the root answer. From there, we sample a hyperedge to determine the relation type and its neighbors. If the operation is projection, we randomly choose a neighboring node as the answer for the subsequent query and recursively sample based on the next operation. The key distinction from ordinary sampling is that the position of the sampled neighbor may differ between sub-queries $a_1$ and $a_2$; for example, in $r_1(e_1,e_2,x)$ and $r2(x,e_3,e_4,y)$, the varaible $x$ can occupy different positions. To accommodate these differences, position information is stored for both backward and forward passes.

\begin{algorithm}
\caption{Ground Query Type}\label{alg:cap}
\begin{algorithmic}
\Require G is KG with arity >= 2.
\Function {SampleQuery}{$T, q$}
\State $T$ is an arbitrary node of the Knowledge graph
\State $q$ is the query structure
\If {$q$.operation is projection}
    \State Sample edge to obtain relation \textit{r} and neighbor set \textit{n}
    \State RelationType $\leftarrow$ \textit{r}
    \State \textit{NextAnswer} $\leftarrow$ random select \textit{n}
    \State NeighborSet $\leftarrow$ $n - \{T,NextAnswer\}$
    \State Store current position of T in previous edge (if any) and current sampled edge
    \State Subquery $\leftarrow$ SampleQuery(\textit{NextAnswer}, $q$)
    \State \Return (T.op, RelationType, NeighborSet, SubQuery)
\ElsIf{$q$.operation is negation}
    \State \Return (T.op, SubQuery)
\ElsIf{$q$.operation is union or intersection}
    OperationResult $\leftarrow$ []
    \For{T.subquery\_structure}                    
        \State OperationResult.append(SampleQuery(T.subquery\_structure, $q$))
    \EndFor
    \State \Return (T.op, OperationResult)
\EndIf
\EndFunction
\end{algorithmic}
\end{algorithm}

\section{Complexity and Runtime Analysis}

\begin{table*}[ht]
    \centering
    \begin{tabular}{lll}
        \hline
        Query Type & Single Pass (LSGT) & Iterative Model (LKHGT) \\
        \hline
        1p & 0.5187 & 0.7516 \\
        2p & 0.5449 & 1.5486 \\
        2i & 0.5901 & 1.7755 \\
        pi & 0.5096 & 2.4583 \\
        ip & 0.5480 & 2.4869 \\
        2in & 0.5375 & 1.8202 \\
        pin & 0.5310 & 2.5221 \\
        pni & 0.5752 & 2.5202 \\
        2u & 0.5634 & 1.7913 \\
        up & 0.5366 & 2.4643 \\
        \hline
    \end{tabular}
    \caption{Query Type Process Time for 1024 Queries per Batch}
    \label{tab:query_time}
\end{table*}

As shown in Table 6, if we consider the inference time for any transformer model to be approximately 0.5 seconds per batch, it is evident that the computational time for the iterative model grows with the number of nodes in the Query Operator Tree.

\noindent Let \( P \) represent the number of projection nodes and \( L \) the number of logical nodes in the query operator tree. The total number of nodes in the query operator tree can be expressed as:
\[
n = P + L,
\]
where \( n \) denotes the total number of nodes. \newline
\textbf{Time Complexity:} 
\begin{itemize}
    \item \textbf{Single Pass Model:} In the single-pass model, only one transformer inference is required for all queries, which can be represented as \( O(1) \).
    \item \textbf{LKHGT Time Complexity:} In the LKHGT framework, the time complexity primarily stems from the number of transformer inferences required. Each query type necessitates processing through the transformer for each node in the operator tree. Therefore, for LKHGT, the time complexity is:
    \[
    O(n).
    \]
\end{itemize}

\noindent\textbf{Space Complexity:} Assume there are a total of 8 token types for both the Logical Encoder and the Projection Encoder. Each token type has a unique set of Query, Key, and Value matrices. Thus, the total number of parameters required for self-attention in LKHGT is:
\[
O(8d^2),
\]
where \( d \) is the dimensionality of the embeddings. In contrast, the native self-attention mechanism has a space complexity of:
\[
O(d^2).
\]
The space complexity for other components of the model remains consistent with the original transformer architecture.

\end{document}